\documentclass[lettersize,journal]{IEEEtran}
\usepackage{amsmath,amsfonts}
\usepackage{amssymb}
\usepackage{algorithm}
\usepackage{algorithmicx}
\usepackage{array}
\usepackage{textcomp}
\usepackage{stfloats}
\usepackage{url}
\usepackage{verbatim}
\usepackage{graphicx}
\usepackage{cite}
\usepackage{ulem}
\usepackage{threeparttable}
\usepackage{newtxmath}
\usepackage{enumitem}
\usepackage{graphicx}
\usepackage{float}
\usepackage{subfigure} 
\usepackage[
pdfauthor={derajan},
pdftitle={How to do this},
pdfstartview=XYZ,
bookmarks=true,
colorlinks=true,
linkcolor=blue,
urlcolor=blue,
citecolor=blue,
pdftex,
bookmarks=true,
linktocpage=true,
hyperindex=true
]{hyperref}

\begin{document}
	\title{Discriminability-Driven Spatial-Channel Selection with Gradient Norm for Drone Signal OOD Detection}
	\author{Chuhan Feng,
		Jing Li,~\IEEEmembership{Member,~IEEE,} Jie Li, Lu Lv,~\IEEEmembership{Member,~IEEE,}
		Fengkui Gong,~\IEEEmembership{Member,~IEEE}
		\thanks{ \textit{(Corresponding author: Jing Li.)}}
		\thanks{Chuhan Feng, Jing Li, Jie Li, and Fengkui Gong are with the State Key Laboratory of Integrated Services Network, Xidian University, Xi'an, Shaanxi 710071, China (e-mail: fch@stu.xidian.edu.cn; jli@xidian.edu.cn; lijie\_372@stu.xidian.edu.cn; fkgong@xidian.edu.cn).}
		\thanks{Lu Lv is with the School of Telecommunications Engineering, Xidian University, Xi'an 710071, China (e-mail: lulv@xidian.edu.cn).}
	}
	\maketitle
	\begin{abstract}
We propose a drone signal out-of-distribution (OOD) detection algorithm based on discriminability-driven spatial-channel selection with a gradient norm. Time-frequency image features are adaptively weighted along both spatial and channel dimensions by quantifying inter-class similarity and variance based on protocol-specific time-frequency characteristics. Subsequently, a gradient-norm metric is introduced to measure perturbation sensitivity for capturing the inherent instability of OOD samples, which is then fused with energy-based scores for joint inference. Simulation results demonstrate that the proposed algorithm provides superior discriminative power and robust performance via SNR and various drone types.

\end{abstract}

\section{Introduction}
\label{sec1}
\IEEEPARstart{W}{ith}  advancements in technology, drones have been extensively applied in fields such as aerial photography, logistics, and rescue. However, complex environments often introduce unpredictable interference, such as Wireless Fidelity (Wi-Fi), Bluetooth, and radar signals. Neural networks tend to misclassify out-of-distribution (OOD) samples as in-distribution (ID) samples, severely reducing the reliability of drone signal detection\textcolor{blue}{\cite{ref7}}. Thus, it is essential to investigate the characteristics of drone signals and identify an effective OOD detection algorithm to improve accuracy.

OOD detection algorithms have been broadly categorized into four main branches: density-based, distance-based, reconstruction-based, and classification-based methods\textcolor{blue}{\cite{ref1}}. Density-based methods leverage probability density estimation to identify deviations from learned ID distributions but suffer from the curse of dimensionality in high-dimensional drone time-frequency image (TFI) feature vectors\textcolor{blue}{\cite{ref2}}. Distance-based methods rely on geometric metrics to measure proximity to ID clusters, yet struggle to distinguish near-OOD drone signals under interference and noise\textcolor{blue}{\cite{ref3}}. Reconstruction-based methods exploit reconstruction errors from autoencoders, yet the inherent information bottleneck often suppresses subtle protocol-specific signatures, yielding representations that lack the discriminative power required for robust detection\textcolor{blue}{\cite{ref10}}. Classification-based methods utilize classifier outputs for OOD detection, where Softmax probabilities often exhibit overconfidence for OOD drone signals\textcolor{blue}{\cite{ref5,tvt_new1}}. 

Conventional channel selection algorithms are primarily developed for natural images, thus overlooking the critical time-frequency distribution characteristics inherent in the spatial dimension of signal TFIs\textcolor{blue}{\cite{ref6}}. To address this, the conversion of in-phase/quadrature (I/Q) samples into TFIs is utilized to overcome the limited discriminative information in raw I/Q samples. Specifically, TFIs explicitly capture protocol-specific time-frequency patterns, thereby effectively revealing distinguishing features in the time-frequency distributions of ID and OOD samples. The time-frequency properties of various signals can be quantified by leveraging the spatial components of TFI feature vectors, since the duration, interval, and frequency-domain distributions specified by communication protocols exhibit distinct characteristics\textcolor{blue}{\cite{tvt_new2}}. Consequently, spatial selection is incorporated to amplify the discriminative time-frequency features of ID samples by adaptively weighting spatial and channel features based on inter-class differences. This weighting mechanism enables the model to assign lower energy-based confidence scores to OOD samples, as they deviate from the learned ID distribution in the enhanced feature space.

Besides, existing methods relying solely on static confidence scores (e.g., energy-based scores) fail to capture the dynamic stability of predictions under environmental perturbations, such as Doppler shifts and co-channel interference. OOD samples near decision boundaries are highly susceptible to minor input variations and yield unstable predictions. To address this limitation, model-prediction sensitivity is introduced as a core discriminative criterion for drone-signal OOD detection, with perturbations applied to model predictions. Specifically, a gradient norm is designed to quantify this sensitivity by measuring the magnitude of gradients backpropagated from classification logits to feature vectors. This dynamic stability indicator is fused with the traditional energy score via weighted averaging to enable more precise OOD detection, particularly for samples with high confidence yet unstable predictions under perturbations.

In this paper, we propose a discriminability-driven spatial-channel selection algorithm with gradient norm for drone signal OOD detection. The main contributions are summarized as follows:
\begin{itemize}
\item{A plug-and-play discriminability-driven spatial-channel selection module is developed. It enhances feature discriminability by adaptively weighting spatial and channel grids based on inter-class statistical disparities, effectively widening the gap between ID and OOD signals.} 


\item{A gradient norm quantifies perturbation sensitivity by measuring logit derivatives relative to feature vectors. Robust OOD detection is achieved by integrating this dynamic indicator with static energy scores for joint inference.} 


\item{Simulation evaluations demonstrate that discriminability-driven spatial-channel selection with the gradient norm (DDSCS) algorithm consistently outperforms existing algorithms across multiple metrics, while exhibiting exceptional robustness via SNR and various drone types.} 
\end{itemize}

\section{Signal Analysis and Algorithm Description}
\label{sec2}
\subsection{Signal Analysis}
In this section, the foundation for signal representation is established based on time-frequency analysis of drone signals.

\begin{enumerate}[leftmargin=0pt, itemindent=2pc, listparindent=\parindent]	
\item{ \textit{TFI Features}: 
	The received drone I/Q sequence $x$ is transformed into TFIs using the short-time Fourier transform (STFT)\textcolor{blue}{\cite{ref8}}, which can be expressed as
	\begin{equation}
		X(t,f)\text{=}\left| \sum\limits_{\tau =-\infty }^{+\infty }{x{{h}^{*}}( \tau-t ){{e}^{-j2\pi f\tau }}} \right|,
	\end{equation}
	which captures the joint time-frequency characteristics. The \(h(\tau-t)\) denotes the window function located at the $t$-th time index, and the window length $\tau$ is equal to the fast Fourier transform size. However, the single-channel $X(t,f)$ is extended to a three-channel tensor $\mathbf{I}\in\mathbb{R}^{3 \times H \times W}$ in the context of deep learning input, where $H$ corresponds to time $T$, $W$ corresponds to frequency $f$. The three channels replicate the STFT magnitude across red, green, and blue channels to produce TFI.
}

\item \textit{Traditional Energy Measurement}: 
Traditional time-frequency analysis is typically performed using energy-based algorithms, where signal appearances are constrained to specific temporal and spectral regions on the TFI by communication protocols. To implement this, the temporal energy vector is first computed by summing the time frequency image components across the frequency axis as
\begin{equation}
	\mathbf{F}_i = \sum_{j=1}^{W} X_{i,j}, \quad i \in {1,\dots,H}, \quad j \in {1,\dots,W}.
\end{equation}

The frequency bin characterized by the maximum energy is then identified by

\begin{equation}
	k = \arg\max_{i} \mathbf{F}_i.
\end{equation}

Based on this identified bin, a mean magnitude response is utilized as the representative energy metric for OOD sample identification, which is calculated as
\begin{equation}
	S{\text{energy}} = \frac{1}{W} \sum_{j=1}^{W} X_{k,j}
	\label{eq:energy_score}.
\end{equation}
\end{enumerate}

However, identifying discriminative regions based solely on such mean magnitude responses is challenging, as complex environments can obscure protocol-specific signatures. Furthermore, traditional analysis based on energy maximization is ineffective at distinguishing fixed-frequency signals from transient interference and remains highly susceptible to noise. The consistent energy distribution of fixed-frequency signals prevents these maximum-energy metrics from adequately capturing temporal frequency disparities, motivating the development of the proposed algorithm.

\subsection{Spatial-Channel Selection}
In this section, a neural network generates features from the TFI and subsequently reweights them along both spatial and channel dimensions to enhance discriminability before feature integration. Spatial-channel components exhibit low inter-class similarity and high variance, and are assigned higher weights as they indicate high discriminability.

The extracted feature vector can be expressed as
\begin{equation}
\mathbf{F}=\phi(\mathbf{I})\in\mathbb{R}^{C_{\text{s}}\times H_{\text{s}}\times W_{\text{s}}},
\end{equation}
where $\phi(\cdot)$ denotes the MobilenetV2 network\textcolor{blue}{\cite{ref9}}. The feature dimensions are set as $C_\text{s}=1280$ channels with spatial dimensions $H_{\text{s}}=W_{\text{s}}=7$ determined by network downsampling.

\subsubsection{Spatial Selection}

The spatial selection assigns higher weights to spatial features with low inter-class similarity and high inter-class variance, thereby selecting highly discriminative features.

The channel-averaged feature vector can be calculated by 
\begin{equation}
\mathbf{M}_{s,n}^{(c)} = \frac{1}{C_{\text{s}}}\sum_{k=1}^{C_{\text{s}}}  \mathbf{F}_n^{(c)}(k,:,:) \in \mathbb{R}^{1\times H\times W},
\end{equation}
where $\mathbf{F}_n^{(c)}(k,:,:)$ represents the $k$-th channel of $\mathbf{F}$ for the $n$-th sample in $c$-th class.

The class-wise spatial mean of the $(i,j)$-th component can be computed by
\begin{equation}
\mu_c^{i,j} = \frac{1}{N_c}\sum_{n=1}^{N_c} \mathbf{M}_{s,n}^{(c)}(1,i,j),
\end{equation}
where $N_c$ is the number of samples in $c$-th class. 

The inter-class similarity of each spatial component can be quantified by
\begin{equation}
S_{i,j}^{\text{spatial}} = \frac{1}{N_{\text{cls}}(N_{\text{cls}}-1)} \sum_{p=1}^{N_{\text{cls}}} \sum_{q=1, q\neq p}^{N_{\text{cls}}} \text{sim}(\mu_p^{i,j}, \mu_q^{i,j}),
\end{equation}
where $N_{\text{cls}}$ is the total number of classes, and sim(·,·) denotes the cosine similarity operation. 

The inter-class variance of each spatial component can be computed by
\begin{equation}
V_{i,j}= \frac{1}{N_{\text{cls}}}\sum_{c=1}^{N_{\text{cls}}} (\mu_c^{i,j} - \frac{1}{N_{\text{cls}}}\sum_{c=1}^{N_{\text{cls}}}\mu_c^{i,j})^2.
\end{equation}

The spatial discriminability is quantified by combining inter-class similarity and variance with $\alpha$, which can be formulated by
\begin{equation}
S_{i,j} =  (1-\alpha) \cdot V_{i,j}- \alpha \cdot S_{i,j}^{\text{spatial}}, \quad \alpha \in [0,1].
\end{equation}

The spatial weights can be computed by
\begin{equation}
W_{i,j}^s = \frac{S_{i,j}}{\sum_{i=1}^{H}\sum_{j=1}^{W} S_{i,j}}.
\end{equation}

The spatially weighted features can be obtained by
\begin{equation}
\mathbf{F}_s = \mathbf{F} \odot \mathbf{W}^s,
\end{equation}
where $\odot$ denotes element-wise multiplication.

\subsubsection{Channel Selection}
Following spatial selection, features are adaptively reweighted along the channel dimension by assessing inter-class similarity and variance. 

For each channel $k \in \{1,2,\ldots,C_{\text{s}}\}$ and $c$-th class, the channel average can be calculated by
\begin{equation}
\bar{f}_{k,c} = \frac{1}{N_c}\sum_{n=1}^{N_c}\frac{1}{H \times W}\sum_{i=1}^{H}\sum_{j=1}^{W} \mathbf{F}_{s,n}^{(c)}(k,i,j).
\end{equation}

The inter-class channel similarity can be computed by
\begin{equation}
S_{k}^{\text{channel}} = \frac{1}{N_{\text{cls}}(N_{\text{cls}}-1)} \sum_{p=1}^{N_{\text{cls}}} \sum_{\substack{q=1 \\ q \neq p}}^{N_{\text{cls}}} \text{sim}(\bar{f}_{p,k}, \bar{f}_{q,k}).
\end{equation}
which measures the average differences between all class pairs for the $k$-th channel. 

The inter-class channel variance can be estimated as
\begin{equation}
V_k = \frac{1}{N_{\text{cls}}}\sum_{c=1}^{N_{\text{cls}}} (\bar{f}_{k,c} - { \frac{1}{N_{\text{cls}}}\sum_{c=1}^{N_{\text{cls}}}\bar{f}_{k,c}})^2.
\end{equation}

The channel discriminability is quantified by combining inter-class similarity and variance with $\beta$, which can be formulated as
\begin{equation}
T_k =  (1-\beta) \cdot V_k - \beta \cdot S_k^{\text{channel}}, \quad \beta \in [0,1],
\end{equation}
where $\beta$ balances the importance of the dissimilarity and variance channels.

The adaptive channel weights can be calculated by
\begin{equation}
W_k^c = \frac{T_k}{\sum_{k=1}^{C_{s}} T_k}.
\end{equation}

The final spatial-channel weighted features can be obtained by
\begin{equation}
\mathbf{F}_{\text{sc}} = \mathbf{F}_s \odot \mathbf{W}^c.
\end{equation}

\subsubsection{Feature Aggregation}
$\mathbf{F}_{\text{sc}}$ are aggregated into a global representation vector $\mathbf{g}$ via global average pooling, which can be obtained by
\begin{equation}
\mathbf{g} = \frac{1}{H \times W}\sum_{i=1}^{H}\sum_{j=1}^{W} \mathbf{F}_{\text{sc}}(:,i,j) \in \mathbb{R}^{C_{s}}.
\end{equation}

This vector $\mathbf{g}$ is then fed into a fully-connected layer to produce the final logits as
\begin{equation}
\mathbf{z} = \mathbf{W}_{\text{sc}}\mathbf{g} + \mathbf{b}, \quad \mathbf{W}_{\text{sc}} \in \mathbb{R}^{ N_{\text{cls}}\times{C_{\text{s}}}},
\end{equation}
where $\mathbf{W}_{\text{sc}}$ and $\mathbf{b}$ are parameters optimized during the network training.

\subsection{Fusion of Gradient Norm and Energy Score}
The model prediction sensitivity to perturbations is quantified by a gradient norm and fused with the static energy measure via the dynamic stability indicator.

\subsubsection{Energy-based Score}
The energy-based score provides a measure of prediction confidence by aggregating information across all logits, which is expressed as
\begin{equation}
S_{\text{energy}} = \log \sum_{j=1}^{N_{\text{cls}}} \exp(z_j).
\end{equation}

\subsubsection{Gradient Norm}
For a given sample with $\mathbf{g} $, the gradient of the classification logits can be computed by
\begin{equation}
\nabla_{\mathbf{g}} \mathbf{z} = \frac{\partial \mathbf{z}}{\partial \mathbf{g}} \in \mathbb{R}^{C_{\text{s}} \times N_\text{cls}},
\end{equation}
where $\mathbf{z}$ represents the logit vector, and $\frac{\partial}{\partial \mathbf{g}}$ indicates partial differentiation.

The gradient norm can be calculated by
\begin{equation}
G_{\text{norm}} = \left\|\nabla_{\mathbf{g}} \max_j z_j\right\|_2 = \sqrt{\sum_{d=1}^{C_{\text{s}}} \left(\frac{\partial \max_j z_j}{\partial g_d}\right)^2},
\end{equation}
where $\|\cdot\|_2$ denotes the L2 norm, $g_d$ is the $d$-th element of $\mathbf{g}$, and $z_j$ is the $j$-th logit. $G_{\text{norm}}$ quantifies the instability of OOD samples, with larger values representing higher sensitivity to minor perturbations.

\subsubsection{Fused Score}
The final OOD detection score combines $S_{\text{energy}}$ with $G_{\text{norm}}$ through weighted fusion, which is computed by
\begin{equation}
S_{\text{fused}} = \lambda \cdot {\cal N}(S_{\text{energy}}) - (1-\lambda) \cdot {\cal N}(G_{\text{norm}}),
\end{equation}
where $\lambda$ is a weighting factor, and $\cal N(\cdot)$ denotes the unified z-score normalization function. 

The threshold $\gamma$ determines whether the input sample belongs to the ID or OOD distribution, which can be expressed as
\begin{equation}
\text{Label}(x) = \begin{cases}
	\text{ID} & \text{if } S_{\text{fused}} \geq \gamma, \\
	\text{OOD} & \text{if } S_{\text{fused}} < \gamma.
\end{cases}
\end{equation}

$\gamma$ is typically selected based on validation set performance to achieve an optimal detection trade-off.

\section{Experiment}
\subsection{Set Up}
In DroneRFa, raw data are obtained by sampling 15 drone types from T0001 to T1111 and background noise designated as T0000, all of which incorporate inherent environmental impairments\textcolor{blue}{\cite{ref4}}. Subsequently, additive white Gaussian noise is added to the background interference, with SNR ranging from -15 dB to 15 dB in 2 dB increments. A total of 31,873 in-phase and quadrature samples generated through this process are evenly distributed across 16 categories and are split into training, validation, and test sets at a ratio of 8:1:1. OOD samples are reserved solely for the test set to evaluate model performance. The hardware environment consists of an NVIDIA GeForce RTX 4060 Ti graphics processing unit and an 11th Gen Intel Core i7-11700K central processing unit. The software environment is composed of PyTorch 1.12.1, Python 3.9, and CUDA 11.6.

Five baseline algorithms are selected for a comprehensive performance comparison, and the efficacy of the proposed OOD detection algorithm is evaluated against established methodologies.

\begin{enumerate}[leftmargin=0pt, itemindent=2pc, listparindent=\parindent]
\item \textit{MobileNetV2-Based}: I/Q samples are directly converted into TFIs, and the native MobileNetV2 architecture is utilized without any OOD detection modules added.

\item \textit{Softmax-Based}: Features extracted by MobileNetV2 are processed by the Softmax to obtain a confidence score based on which OOD detection is performed. 

\item \textit{Uncertainty-Based}: A confidence estimation branch is integrated into the MobileNetV2 backbone to compute a confidence score. Samples are classified as ID when their confidence scores exceed an optimized threshold.

\item \textit{Variational Autoencoders-Based}: A Variational Autoencoder is trained to model the distribution of ID samples with OOD detection based on the reconstruction error\textcolor{blue}{\cite{ref11}}. 

\item \textit{Discriminability-driven channel selection (DDCS)}: Features extracted by MobileNetV2 undergo a channel selection process to increase their discriminability, and the parameters are re-optimized on the current dataset\textcolor{blue}{\cite{DDCS}}. 

\end{enumerate}

\subsection{Simulation Results}

Four widely recognized metrics were utilized to evaluate model performance comprehensively. Accuracy provides an overall measure of classification correctness, while Recall focuses on the model's ability to identify positive instances. The F1 score, as the harmonic mean of precision and Recall, offers a balanced assessment of model performance. The Area Under the Receiver Operating Characteristic Curve (AUROC) measures the overall discriminative power across different classification thresholds, with higher values indicating greater ability to distinguish between ID and OOD samples. Weighted evaluation metric is defined as $\text{WEM} = \frac{1}{4} \times (\text{Accuracy} + \text{F1} + \text{Recall} + \text{AUROC})$.

\begin{figure}[!t]
\centering
\includegraphics[width=2.5in]{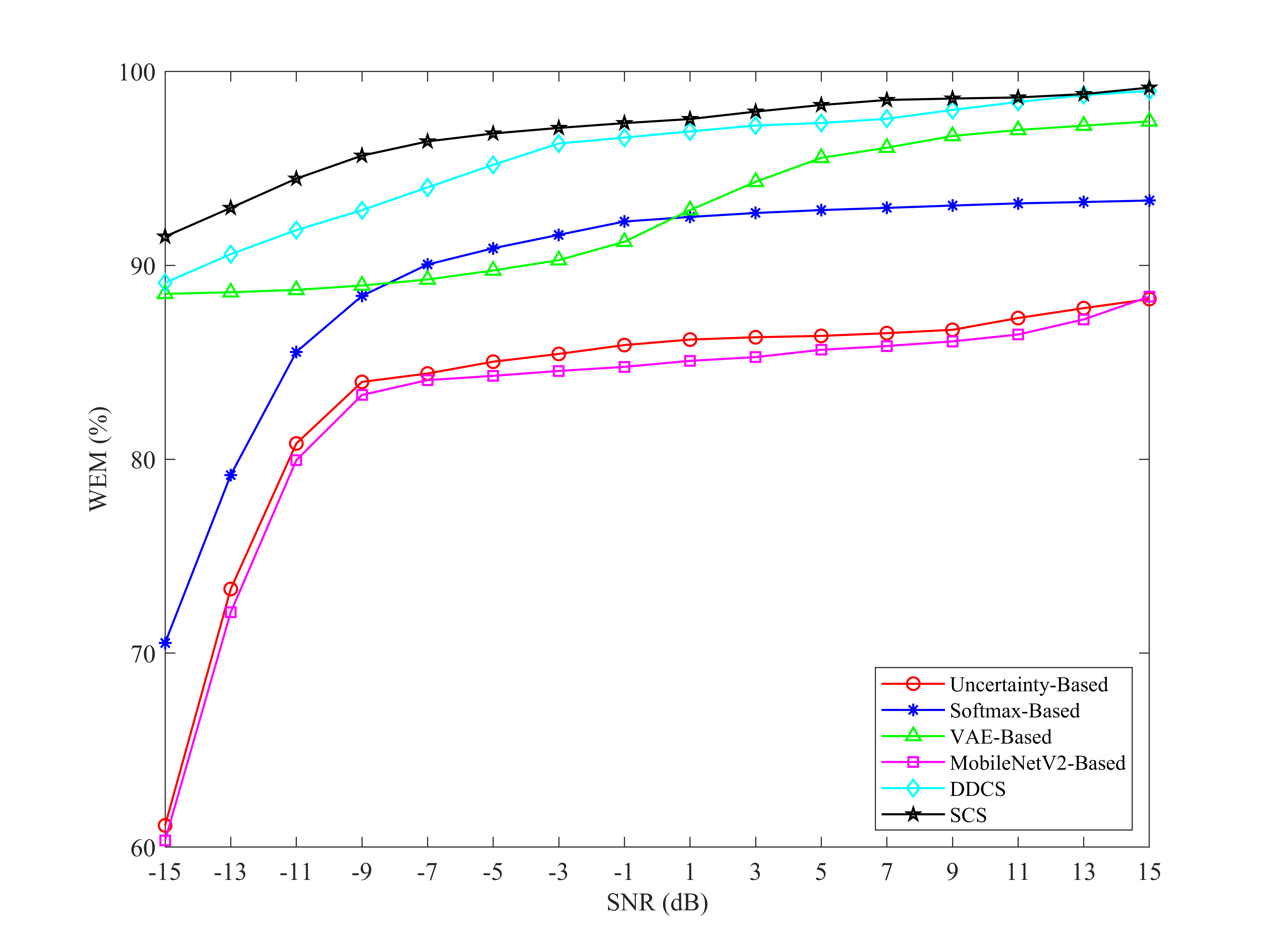}
\caption{WEM performance of various algorithms via SNR.}
\label{Fig_1}
\end{figure}

\begin{figure}[!t]
\centering
\includegraphics[width=2.5in]{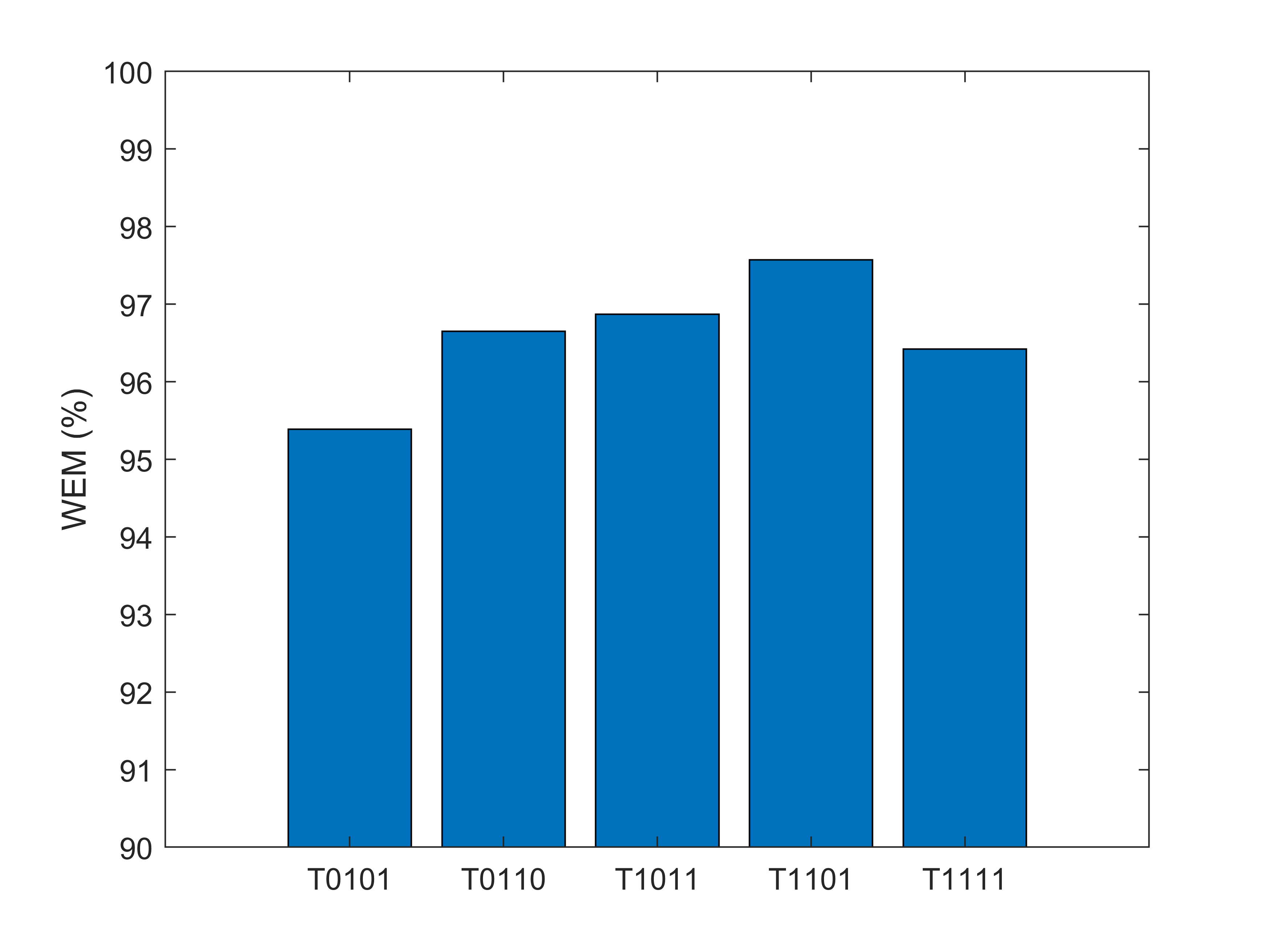}
\caption{WEM of the proposed algorithm versus various drone types.}
\label{Fig_2}
\end{figure}

Fig. \ref {Fig_1} illustrates the SNR-resilience of the evaluated algorithms. DDSCS maintains a substantial performance margin, particularly in low SNR regimes from -15 dB to -5 dB, as its spatial-channel selection effectively extracts protocol-specific TF patterns while mitigating noise-induced artifacts. The robustness across heterogeneous OOD types is validated in Fig. \ref{Fig_2}, where T0101, T0110, T1011, T1101, and T1111 are respectively designated as OOD samples. The proposed algorithm achieves a minimum WEM of 95.43\%, confirming its ability to learn generalized discriminative boundaries between ID protocols and external interference rather than merely memorizing signal instances.

Quantitatively, Table \ref{tab:acc} shows that DDSCS significantly outperforms the second-best baseline. By achieving 95.18\% accuracy and 95.77\% AUROC, the proposed algorithm demonstrates superior capability in identifying unknown drone signals compared to traditional classification and reconstruction-based methodologies.

\begin{table}[!t]
\caption{Evaluation Metrics of Different Algorithms\label{tab:acc}}
\centering
\begin{tabular}{ccccc}
	\hline
	Algorithm &Accuracy (\%) & Recall (\%)  & F1 (\%) & AUROC (\%) \\
	\hline
	Uncertainty-Based & 84.63&93.26& 87.51 & 84.55  \\
	Softmax-Based & 90.86  &93.82& 87.74&91.81\\
	VAE-Based & 88.82 &  97.39  & 93.50 & 83.17\\
	MobileNetV2-Based & 82.40 &92.06&  86.55& 80.76  \\
	DDCS & 93.35 & 97.05 & 95.60  & 92.40  \\
	DDSCS & \textbf{95.18}   &\textbf{98.65}  &  \textbf{96.42}& \textbf{95.77}  \\
	\hline
\end{tabular}
\end{table}

\subsection{Ablation Study}
To evaluate each component's contribution, we conduct an ablation study on the following variants. 

\begin{itemize}
\item \textit{Spatial-only}: Selection based solely on spatial dimensions.
\item \textit{Channel-only}: Selection based solely on channel dimensions.
\item \textit{Spatial-Channel}: Joint spatial-channel selection without gradient norm.
\item \textit{DDSCS}: Full model with spatial-channel selection and gradient norm fusion.
\end{itemize}

Table \ref{tab:accuracy} quantifies the contribution of each module to the overall performance. Among the individual components, spatial selection is identified as the most impactful driver, significantly enhancing the discriminative power of protocol-specific time-frequency features. The integration of channel selection further yields synergistic improvements by amplifying distinctive spectral signatures.

The gradient norm provides the final performance leap, achieving peak detection accuracy. This result confirms that quantifying prediction sensitivity to perturbations is more effective for OOD detection than relying solely on static features. By penalizing overconfident yet unstable predictions, the proposed DDSCS algorithm successfully captures the dynamic instability of OOD samples that is typically missed by conventional confidence scores.

\begin{table}[!t]
\caption{Evaluation Metrics of Different Models\label{tab:accuracy}}
\centering
\begin{tabular}{ccccc}
	\hline
	Algorithm &Accuracy (\%) & Recall (\%)  & F1 (\%) & AUROC (\%) \\
	\hline
	Spatial-only & 94.15 &  97.39  & 95.83 & 93.17\\
	Channel-only & 93.35  &97.05&  95.60& 92.40  \\
	Spatial-Channel &95.07 & 98.26 & 96.13 & 94.56  \\
	DDSCS & \textbf{95.18}   &\textbf{98.65}  &  \textbf{96.42}& \textbf{95.77}  \\
	\hline
\end{tabular}
\end{table}

\subsection{Sensitivity Analysis}

\subsubsection{Hyperparameters $\alpha$ and $\beta$}
\begin{figure}[!t]
	\centering
	\includegraphics[width=2.8in]{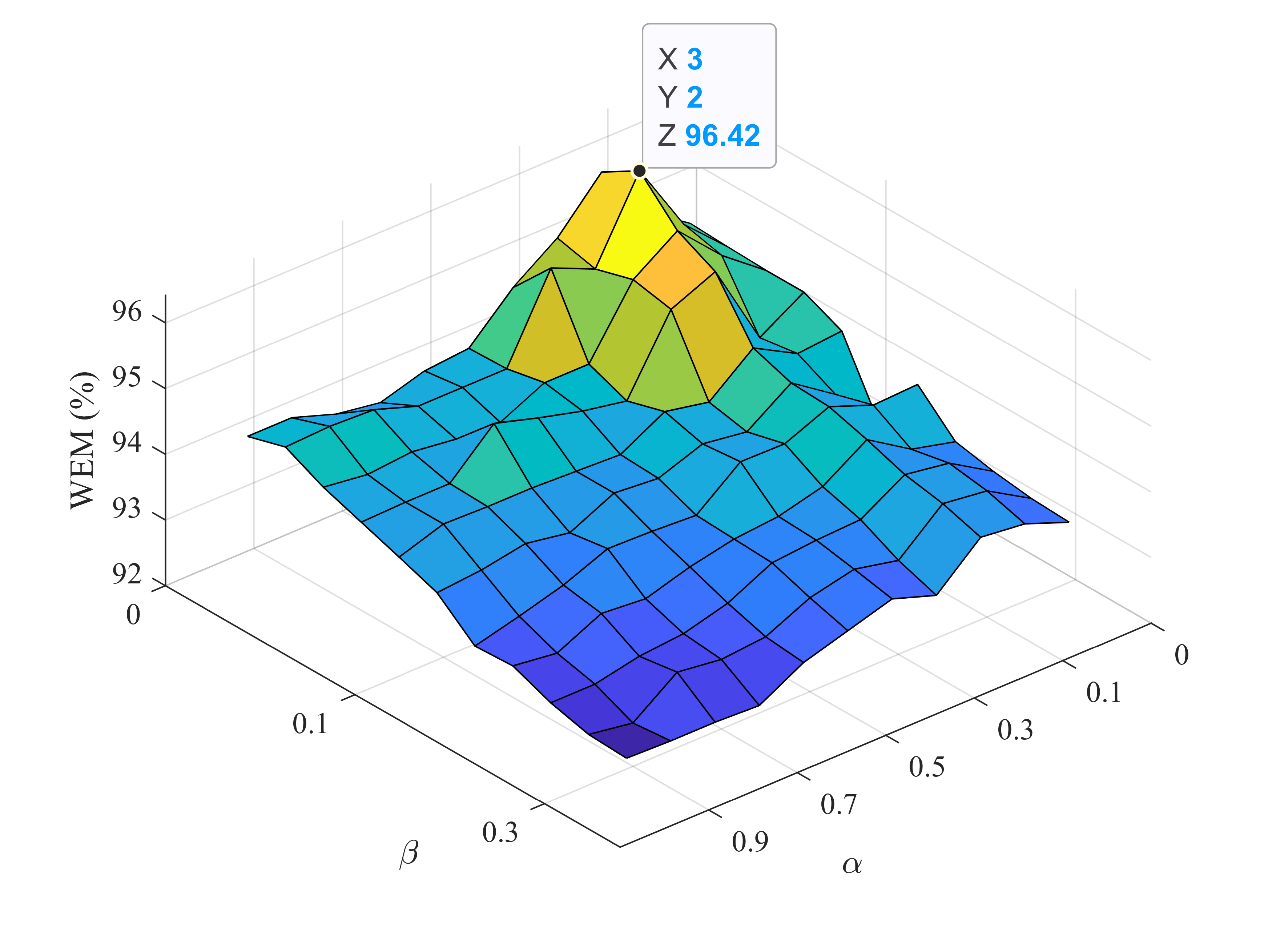}
	\caption{Sensitivity analysis of selection parameters $\alpha$ and $\beta$.}
	\label{Fig_3}
\end{figure}

As shown in Fig. \ref{Fig_3}, peak accuracy is achieved at $\alpha = 0.1$ and $\beta = 0.2$. The preference for lower $\alpha$ and $\beta$ values indicates that the model benefits more from emphasizing inter-class variance rather than merely minimizing similarity. The result suggests that the most discriminative features are those with highly diverse activation signatures across different communication protocols for RF signals.

\subsubsection{Fusion Weight $\lambda$}
\begin{figure}[!t]
	\centering
	\includegraphics[width=2.8in]{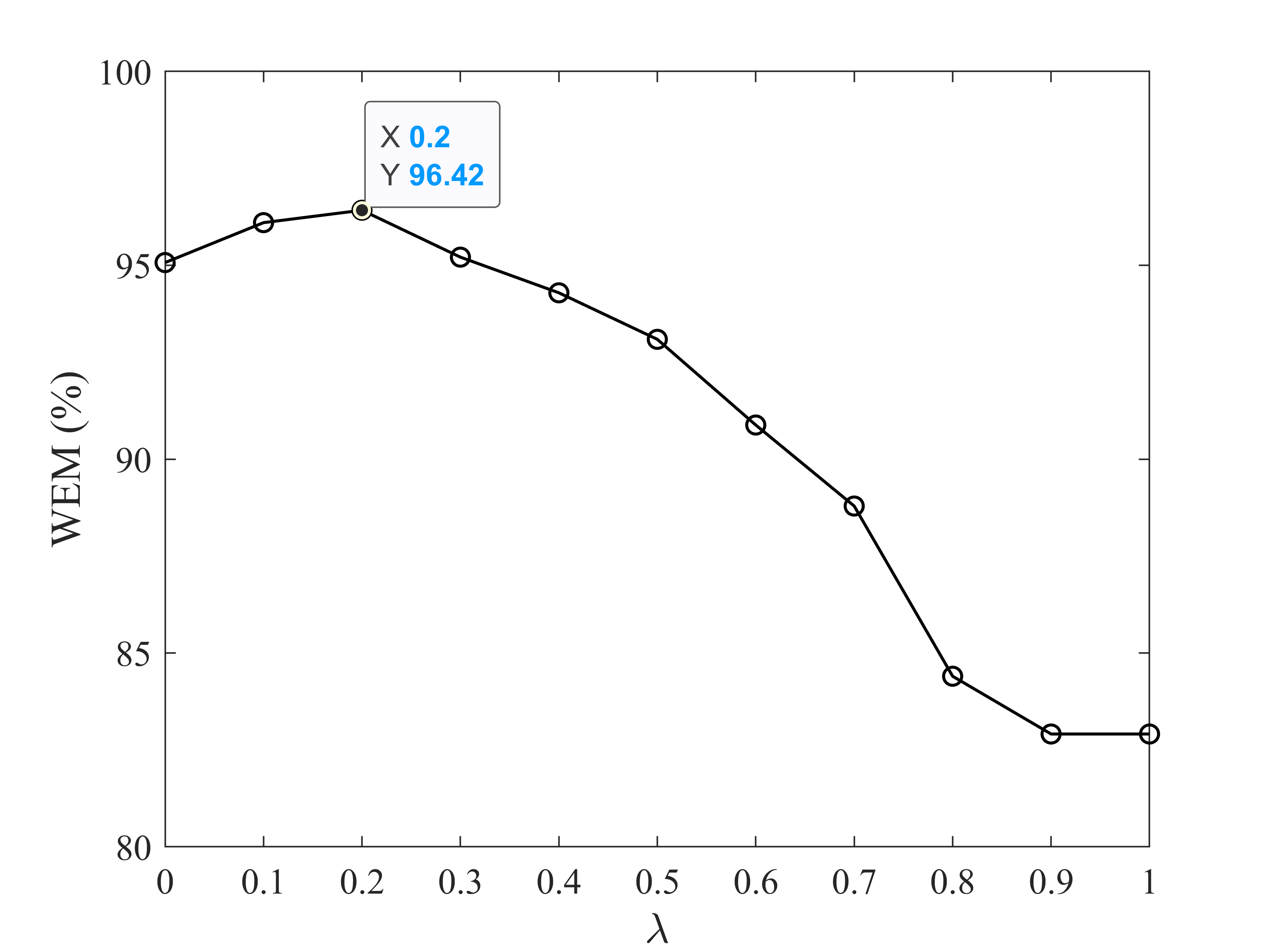}
	\caption{Impact of fusion weight $\lambda$ on detection performance.}
	\label{Fig_4}
\end{figure}

The sensitivity of the OOD score fusion is analyzed in Fig. \ref{Fig_4}. Performance degrades at extreme values of $\lambda$. A slight $\lambda$ over-amplifies gradient noise, leading to false OOD alarms, while a large $\lambda$ ignores the instability of OOD samples, failing to detect "near-boundary" threats. The optimal value $\lambda = 0.2$ reflects a balance where the energy score provides the fundamental confidence measure, while the gradient norm provides essential stability-based refinement.

\section{Conclusion}

This paper proposed the discriminability-driven spatial-channel selection with gradient norm for drone signal OOD detection. A discriminability-driven spatial-channel selection module was introduced, which applied the adaptive weight assignment for spatial and channel components based on time-frequency distribution characteristics. Furthermore, a gradient norm was designed to quantify prediction sensitivity to perturbations and was effectively combined with the energy-based score via a weighted average. Experimental results demonstrated superior performance across multiple metrics, with Accuracy, Recall, F1, and AUC improved by 1.83\%, 1.26\%, 0.82\%, and 3.37\%, respectively. The proposed algorithm also exhibited robust performance via SNR and various drone types.

\end{document}